\documentclass[fleqn,letter,12pt]{report}
\usepackage{amsmath}
\usepackage{latexsym}
\usepackage{amsfonts}
\usepackage{amssymb}
\usepackage{amsthm}
\usepackage{indentfirst}
\usepackage{empheq}
\usepackage{algorithmic}
\usepackage{titlesec}

\titleformat{\paragraph}
{\normalfont\normalsize\bfseries}{\theparagraph}{1em}{}
\titlespacing*{\paragraph}
{0pt}{3.25ex plus 1ex minus .2ex}{1.5ex plus .2ex}

\begin{document}

\title{An Unsupervised, Iterative $N$-Dimensional Point-Set Registration Algorithm}

\author{A. Pasha Hosseinbor \and R. Zhdavov \and A. Ushveridze}

\maketitle

\begin{abstract}
An unsupervised, iterative $N$-dimensional point-set registration algorithm for unlabeled data (i.e. correspondence between points is unknown) and based on linear least squares is proposed. The algorithm considers all possible point pairings and iteratively aligns the two sets until the number of point pairs does not exceed the maximum number of allowable one-to-one pairings. 
\end{abstract}


\section{Introduction}
Point-set registration has been extensively studied in computer vision. For labeled data, an important class of solutions is linear least squared techniques \cite{kabsch.1976,arun.1987,umeyama.1991,chang.1997}. Unlike the aforementioned work of \cite{kabsch.1976,arun.1987,umeyama.1991}, the paper in \cite{chang.1997} treated the more general case of two point-sets of unequal size and do not assume correspondence. They first established correspondence by numerically determining the matching pairs support between the two points sets (i.e. finding an optimal subset of pairings between the two sets), and then derived (analytical) least squared solutions to the transformation parameters that optimally align the (labeled) optimal subset of pairings. However, their approach is computationally expensive, having quartic polynomial (average-case) complexity.

For unlabeled data, the alignment of two point patterns is a two-part problem; both the correspondence and the optimal affine transform that minimizes some dissimilarity metric between the two point-sets need to be determined. Various point pattern matching algorithms for unlabeled data have been proposed in \cite{rangarajan.1997,rangarajan2.1997,gold.1998,tsin.2004,jian.2010}, but the necessary optimization for each is numerically based. The works of \cite{rangarajan.1997,rangarajan2.1997,gold.1998} determined the optimal affine transform and correspondence simultaneously by numerically solving a constrained least squares problem. The method in \cite{jian.2010} models each point-set as a Gaussian mixture, and determines the appropriate alignment by (non-linear) optimization of the L2 distance between the two Gaussian mixtures. 

Unlike the numerical optimization schemes discussed above, a much more computationally efficient approach would be an analytical optimization scheme for unlabeled data, which is proposed in this paper. Our alignment approach is similar to the labeled techniques of \cite{kabsch.1976,arun.1987,umeyama.1991,chang.1997} in that we derive analytical solutions for the optimal affine transform via linear least squares. But unlike them, we do not assume or establish correspondence prior to registration, but use registration to establish correspondence. To the best of our knowledge, our derivation of the (closed-form) linear least square solutions for the registration of two unlabeled point-sets, though remarkably simple, is absent from the available literature. The presented derivation is generalized to any $N$-dimensional point-set (i.e. each point in a point-set resides in $\mathbb{R}^{N}$), and the obtained solutions are then used to create an unsupervised, iterative $N$-dimensional point-set registration algorithm.  The $N=2$ case was shown in \cite{hosseinbor.2017}, and utilized in the context of fingerprint matching.

The paper is organized as follows. In Section II, we lay the theoretical foundations of the proposed algorithm; in Section III, we describe its numerical implementation; and in Section IV, we conclude with a discussion of the algorithm and its potential applications. 

\section{Theory}
Consider two $N$-dimensional point sets ${\bf U}$ and ${\bf V}$ comprising $N_{U}$ and $N_{V}$ singular points, respectively. The Cartesian coordinates of the singular points will be expressed as an $N$-dimensional vector:

\begin{align*}
{\bf u}_{i} \in \mathbb{R}^{N} \in {\bf U} \; \; \; i=1,\dots,N_{U} \\
{\bf v}_{k} \in \mathbb{R}^{N} \in {\bf V} \; \; \; k=1,\dots,N_{V} 
\end{align*}
The elements of ${\bf u}_{i}$ and ${\bf v}_{k}$ are denoted as $u_{i}^{j}$ and $v_{k}^{j}$, respectively, where $j=1,..., N$. ${\bf U}$ and ${\bf V}$ can be interpreted as matrices whose columns are formed by the vectors ${\bf u}_{i}$ and ${\bf v}_{k}$, respectively, i.e. ${\bf U} \in \mathbb{R}^{N_{U} \times N}$ and ${\bf V} \in \mathbb{R}^{N_{V} \times N}$; and $u^{j}$ and $v^{j}$ can be interpreted as the features of ${\bf U}$ and ${\bf V}$, respectively.

We want to register point-set ${\bf U}$ to ${\bf V}$. There are $N_{U}N_{V}$ possible matching (cross) pairs and at most $\min\{N_{U},N_{V}\}$ one-to-one matching pairs. Let $m_{ik}$ denote the weight of a matching pair; the weight can be interpreted as a probability that the points ${\bf u}_{i}$ and ${\bf v}_{k}$ match locally. We assume that the (initial) $m_{ik}$ for each (cross) pair has been determined prior to alignment. One way of computing $m_{ik}$ is discussed in \cite{hosseinbor.2017} within the context of fingerprint matching. 

\subsection{Case I: No Scale}

We apply a global rotation and translation to point set ${\bf U}$ such that 

\begin{equation}
{\bf u}_{i}^{'} = {\bf L}{\bf u}_{i} + \bf{t},
\end{equation}
where $L$ is the $N \times N$ rotation matrix and ${\bf t}$ is the $N$-d vector of translation parameters. Since ${\bf L}$ is a rotation matrix, it is orthogonal, i.e. ${\bf L}{\bf L}^{T} = {\bf L}^{T}{\bf L} = {\bf I}_{N \times N}$, and has a determinant of 1. 

The measure of closeness of the transformed point set ${\bf U}$ and the template set ${\bf V}$ may be defined as the weighted sum of the squared distances between their points (i.e. Euclidean distance metric):

\begin{equation}
e({\bf U},{\bf V};{\bf L},{\bf t} )=\frac{\sum_{i=1}^{N_{U}}\sum_{k=1}^{N_{V}}m_{ik}({\bf u}_{i}^{'}-{\bf v}_{k})^{T}({\bf u}_{i}^{'}-{\bf v}_{k})}{\sum_{i=1}^{N_{U}}\sum_{k=1}^{N_{V}}m_{ik}}
\end{equation}
Interpreting $m_{ik}$ as the probability of a match between ${\bf u}_{i}$ and ${\bf v}_{k}$ leads to $\sum_{i=1}^{N_{U}}\sum_{k=1}^{N_{V}}m_{ik}=1$. Expanding out the product yields

\begin{multline}
\label{eq:dist}
e({\bf U},{\bf V};{\bf L},{\bf t} ) = \sum_{i=1}^{N_{U}}\sum_{k=1}^{N_{V}}m_{ik}[ {\bf v}_{k}^{T}{\bf v}_{k} - 2 \text{Tr}\left({\bf L}{\bf u}_{i}{\bf v}_{k}^{T} \right) + 2\text{Tr}\left({\bf L}{\bf u}_{i}{\bf t}^{T} \right) \\
+ {\bf u}_{i}^{T}{\bf u}_{i} - 2{\bf t}^{T}{\bf v}_{k} + {\bf t}^{T}{\bf t} ],
\end{multline}
where 'Tr' denotes the trace operator. We seek the optimal values of parameters ${\bf L}$ and ${\bf t}$ that minimize $e({\bf U},{\bf V};{\bf L},{\bf t} )$, so the constrained optimization problem to be solved is

\begin{equation*}
\begin{aligned}
& \underset{ {\bf L}, {\bf t} }{\text{minimize}}
& & e({\bf U},{\bf V};{\bf L},{\bf t} ) \\
& \text{subject to}
& & {\bf L}{\bf L}^{T} = {\bf I}_{N \times N}, \\
&&& \text{det} \; {\bf L} = 1.
\end{aligned}
\end{equation*}

The Lagrangian for the above (constrained) optimization problem is 

\begin{equation}
\label{eq:Lagrange}
\mathcal{L}({\bf L}, {\bf t}, \boldsymbol{\alpha}, \lambda) = e({\bf U},{\bf V};{\bf L},{\bf t}) + \text{Tr} \left( \boldsymbol{\alpha}({\bf L}{\bf L}^{T} - {\bf I}_{N \times N}) \right) + \lambda \left( \text{det} \; {\bf L} - 1 \right),
\end{equation}
where $\boldsymbol{\alpha}$ is an $N \times N$ symmetric matrix of Lagrangian multipliers (it is symmetric because ${\bf L}{\bf L}^{T}$ is symmetric, and consequently contains $N(N+1)/2$ Lagrangian multipliers), and $\lambda$ is a scalar Lagrangian multiplier. 

Before proceeding further, let us define the weighted average coordinate vectors as 

\begin{eqnarray*}
\overline{\boldsymbol{u}}=\sum_{i=1}^{N_{U}}\sum_{k=1}^{N_{V}}m_{ik}{\bf u}_{i} \\
\overline{\boldsymbol{v}}=\sum_{i=1}^{N_{U}}\sum_{k=1}^{N_{V}}m_{ik}{\bf v}_{k}
\end{eqnarray*}

Optimizing Eq. (\ref{eq:Lagrange}) with respect to ${\bf t}$ yields 

\begin{equation}
\label{eq:trans}
\hat{\boldsymbol{t}} = \overline{\boldsymbol{v}} - \boldsymbol{L}\overline{\boldsymbol{u}}
\end{equation}
Substituting Eq.(\ref{eq:trans}) back into our Lagrangian and then partial differentiating it with respect to ${\bf L}$ yields 

\begin{equation}
\label{eq:pdiff_rot}
\frac{\partial \mathcal{L}}{\partial {\bf L}} = 2\left( \overline{\boldsymbol{v}} \; \overline{\boldsymbol{u}}^{T} - \sum_{i=1}^{N_{U}}\sum_{k=1}^{N_{V}}m_{ik}{\bf v}_{k}{\bf u}_{i}^{T} + \boldsymbol{\alpha L} \right) + \lambda {\bf L}= \boldsymbol{0}_{N \times N}, 
\end{equation}
where $\boldsymbol{0}_{N \times N}$ denotes an $N \times N$ matrix of zeros. Let 

\begin{equation*}
{\bf Z} = \sum_{i=1}^{N_{U}}\sum_{k=1}^{N_{V}}m_{ik}{\bf v}_{k}{\bf u}_{i}^{T} - \overline{\boldsymbol{v}} \; \overline{\boldsymbol{u}}^{T}, 
\end{equation*}
which is a ($N \times N$) cross-covariance matrix. Specifically,

$$
 {\bf Z}=
\begin{pmatrix}
 \text{cov}(u^{1}, v^{1}) &  \text{cov}(u^{1}, v^{2})  &  \text{cov}(u^{1}, v^{3}) & \cdots &  \text{cov}(u^{1}, v^{N}) \\ \vdots & \vdots & \vdots & \vdots & \vdots \\
  \text{cov}(u^{N}, v^{1})  & \text{cov}(u^{N}, v^{2}) & \text{cov}(u^{N}, v^{3}) & \cdots & \text{cov}(u^{N}, v^{N}) 
\end{pmatrix},
$$
where $\text{cov}(u^{j}, v^{j'})$ denotes the covariance between features $u^{j}$ and $v^{j'}$:

\begin{equation}
\text{cov}(u^{j}, v^{j'}) = \sum_{i=1}^{N_{U}}\sum_{k=1}^{N_{V}}m_{ik} u_{i}^{j} v_{k}^{j'} - \left( \sum_{i=1}^{N_{U}}\sum_{k=1}^{N_{V}}m_{ik} u_{i}^{j} \right) \left( \sum_{i=1}^{N_{U}}\sum_{k=1}^{N_{V}}m_{ik} v_{k}^{j'} \right)
\end{equation}
Solving Eq. (\ref{eq:pdiff_rot}) for ${\bf L}$ yields

\begin{equation*}
\hat{\boldsymbol{L}} = \left( \boldsymbol{\alpha} + \frac{\lambda}{2}{\bf I}_{N \times N} \right)^{-1}{\bf Z}
\end{equation*}
Since ${\bf L}$ is an orthogonal matrix, we have

\begin{equation*}
{\bf L}{\bf L}^{T} = \left( \boldsymbol{\alpha} + \frac{\lambda}{2}{\bf I}_{N \times N} \right)^{-1} \boldsymbol{Z}\boldsymbol{Z}^{T} \left( \boldsymbol{\alpha} + \frac{\lambda}{2}{\bf I}_{N \times N} \right)^{-1}  = {\bf I}_{N \times N},
\end{equation*}
which upon isolating the cross-covariance, ${\bf Z}$, yields $\left( \boldsymbol{\alpha} + \frac{\lambda}{2}{\bf I}_{N \times N} \right)^{2} = \boldsymbol{Z}\boldsymbol{Z}^{T}$, so

\begin{equation}
\label{eq:orthog}
\hat{\boldsymbol{L}} = \pm \left( \sqrt{ \boldsymbol{Z}\boldsymbol{Z}^{T} } \right)^{-1} \boldsymbol{Z}
\end{equation}
The minimum of the error function corresponds to $\left( \sqrt{ \boldsymbol{Z}\boldsymbol{Z}^{T} } \right)^{-1} \boldsymbol{Z}$, so we discard the solution $-\left( \sqrt{ \boldsymbol{Z}\boldsymbol{Z}^{T} } \right)^{-1} \boldsymbol{Z}$.

The matrix $\boldsymbol{Z}\boldsymbol{Z}^{T}$ exhibits several interesting properties:

\begin{enumerate}

\item It is real symmetric. It is real because the elements of $\boldsymbol{Z}$ are covariances, which are always real by definition of covariance. And it is symmetric because $\boldsymbol{Z}\boldsymbol{Z}^{T} = \left(\boldsymbol{Z}\boldsymbol{Z}^{T} \right)^{T}$.

\item It is diagonalizable as a consequence of being real symmetric. We will exploit this fact later. 

\item It is positive-definite because ${\bf x}^{T}\boldsymbol{Z}\boldsymbol{Z}^{T}{\bf x} = (\boldsymbol{Z}^{T}{\bf x})^{T}\boldsymbol{Z}^{T}{\bf x} = || \boldsymbol{Z}^{T}{\bf x} ||^{2} > 0$ for any ${\bf x} \in \mathbb{R}^{N}$.

\item It's eigenvalues are all positive as a consequence of positive-definiteness. 

\end{enumerate}

Eq. (\ref{eq:orthog}) requires taking the square root of the matrix $\boldsymbol{Z}\boldsymbol{Z}^{T}$. According to the spectral theorem, any real symmetric matrix can be diagonalized by an orthogonal matrix:

\begin{equation}
\boldsymbol{Z}\boldsymbol{Z}^{T}  = {\bf P} {\bf D} {\bf P}^{T},
\end{equation}
where ${\bf P}$ is an $N \times N$ orthogonal matrix and ${\bf D}$ is the $N \times N$ diagonal matrix; ${\bf P}$ and ${\bf D}$ are formed by the eigenvectors and eigenvalues, respectively, of $\boldsymbol{Z}\boldsymbol{Z}^{T}$.The square root of $ \boldsymbol{Z}\boldsymbol{Z}^{T} $ is then ${\bf P} {\bf D}^{\frac{1}{2}} {\bf P}^{T}$ because 

\begin{equation*}
\left( {\bf P} {\bf D}^{\frac{1}{2}} {\bf P}^{T} \right)^{2} =  {\bf P}{\bf D}^{\frac{1}{2}} {\bf P}^{T} {\bf P} {\bf D}^{\frac{1}{2}} {\bf P}^{T} = {\bf P} {\bf D} {\bf P}^{T} = \boldsymbol{Z}\boldsymbol{Z}^{T} 
\end{equation*}

Now, any $N \times N$ matrix with $N$ distinct eigenvalues has $2^{N}$ square roots because the square root of each eigenvalue can be either positive or negative. If such a matrix is further positive-definite, then it has precisely one positive-definite square root; the positive-definite square root corresponds to the case where only the positive square root of each eigenvalue is taken. We exploit these properties of $\boldsymbol{Z}\boldsymbol{Z}^{T}$ to extract only its positive-definite square-root, so ${\bf D}^{\frac{1}{2}}$ is formed by the positive square roots of the eigenvalues of $\boldsymbol{Z}\boldsymbol{Z}^{T}$. To this end, we employ the notation $\left(\sqrt{\text{arg}}\right)_{\textit{PD}}$ to refer to the positive-definite square root of any square matrix. Taking all this into account, Eq. (\ref{eq:orthog}) can be expressed as

\begin{equation}
\label{eq:orthog_diag}
\hat{\boldsymbol{L}} = {\bf P} \left(\sqrt{{\bf D}}\right)_{\textit{PD}}^{-1} {\bf P}^{T} \boldsymbol{Z}
\end{equation}

In summary, the optimal alignment parameters are

\begin{empheq}[box=\fbox]{align}
\hat{\boldsymbol{L}} = \left( \sqrt{ \boldsymbol{Z}\boldsymbol{Z}^{T} } \right)_{\textit{PD}}^{-1} \boldsymbol{Z} \nonumber \\
\hat{\boldsymbol{t}} = \overline{\boldsymbol{v}} - \left( \sqrt{ \boldsymbol{Z}\boldsymbol{Z}^{T} } \right)_{\textit{PD}}^{-1} \boldsymbol{Z} \; \overline{\boldsymbol{u}} \nonumber 
\end{empheq}
Or equivalently, 

\begin{empheq}[box=\fbox]{align}
\hat{\boldsymbol{L}} = {\bf P} \left(\sqrt{{\bf D}}\right)_{\textit{PD}}^{-1} {\bf P}^{T} \boldsymbol{Z} \nonumber \\
\hat{\boldsymbol{t}} = \overline{\boldsymbol{v}} - {\bf P} \left(\sqrt{{\bf D}}\right)_{\textit{PD}}^{-1} {\bf P}^{T} \boldsymbol{Z} \; \overline{\boldsymbol{u}} \nonumber 
\end{empheq}

The minimum squared error is found by substituting the optimal alignment parameters back into Eq. (\ref{eq:dist}), which yields

\begin{multline*}
e_{\text{min}}  =  \left( \sum_{i=1}^{N_{U}}\sum_{k=1}^{N_{V}}m_{ik}{\bf u}_{i}^{T} {\bf u}_{i} -  \overline{\boldsymbol{u}}^{T} \; \overline{\boldsymbol{u}} \right) + \left( \sum_{i=1}^{N_{U}}\sum_{k=1}^{N_{V}}m_{ik}{\bf v}_{i}^{T} {\bf v}_{i} -  \overline{\boldsymbol{v}}^{T} \; \overline{\boldsymbol{v}} \right) \\
- 2 \text{Tr} \left( \left(\sqrt{\boldsymbol{Z}\boldsymbol{Z}^{T}}\right)_{\textit{PD}} \; \right)
\end{multline*}
Note that 

\begin{equation}
\sum_{i=1}^{N_{U}}\sum_{k=1}^{N_{V}}m_{ik}{\bf u}_{i}^{T} {\bf u}_{i} -  \overline{\boldsymbol{u}}^{T} \; \overline{\boldsymbol{u}} = \sum_{j=1}^{N} \sigma_{u^{j}}^{2},
\end{equation}
where 

\begin{equation*}
\sigma_{u^{j}}^{2} = \sum_{i=1}^{N_{U}}\sum_{k=1}^{N_{V}}m_{ik}\left(u_{i}^{j} \right)^{2} - \left( \sum_{i=1}^{N_{U}}\sum_{k=1}^{N_{V}}m_{ik} u_{i}^{j} \right)^{2}
\end{equation*}
In other words, $\sigma_{u^{j}}^{2}$ is the variance of feature $u^{j}$, and likewise $\sigma_{v^{j}}^{2}$ is the variance of feature $v^{j}$. So we can rewrite the minimum squared error as

\begin{equation}
\begin{split}
e_{\text{min}}  & = \sum_{j=1}^{N} \left( \sigma_{u^{j}}^{2} + \sigma_{v^{j}}^{2} \right) - 2 \text{Tr} \left( \left(\sqrt{\boldsymbol{Z}\boldsymbol{Z}^{T}}\right)_{\textit{PD}} \; \right) \\ 
& = \sum_{j=1}^{N} \left( \sigma_{u^{j}}^{2} + \sigma_{v^{j}}^{2} \right) - 2  \text{Tr} \left( {\bf P} \left(\sqrt{{\bf D}}\right)_{\textit{PD}} {\bf P}^{T} \right)
\end{split}
\end{equation}

Note that had we used $\hat{\boldsymbol{L}}  = -\left( \sqrt{ \boldsymbol{Z}\boldsymbol{Z}^{T} } \right)_{\textit{PD}}^{-1} \boldsymbol{Z}$, the generated error would be 

\begin{equation*}
e'  = \sum_{j=1}^{N} \left( \sigma_{u^{j}}^{2} + \sigma_{v^{j}}^{2} \right) + 2 \text{Tr} \left( \left(\sqrt{\boldsymbol{Z}\boldsymbol{Z}^{T}}\right)_{\textit{PD}} \; \right)
\end{equation*}
The eigenvalues of a positive-definite matrix all always positive, which means $ \text{Tr} \left( \left(\sqrt{\boldsymbol{Z}\boldsymbol{Z}^{T}}\right)_{\textit{PD}} \; \right) > 0$, so $e' > e_{\text{min}}$.


\subsection{Case II: Uniform Scale}
If we include a uniform scale, $s$, into our affine transform, we then have

\begin{equation}
{\bf u}_{i}^{'} = s {\bf L}{\bf u}_{i} + \bf{t}
\end{equation}
So our cost function becomes

\begin{multline}
\label{eq:dist_s}
e({\bf U},{\bf V};{\bf L},{\bf t},s ) = \sum_{i=1}^{N_{U}}\sum_{k=1}^{N_{V}}m_{ik} [ {\bf v}_{k}^{T}{\bf v}_{k} - 2s \; \text{Tr}\left({\bf L}{\bf u}_{i}{\bf v}_{k}^{T} \right) + 2s \; \text{Tr}\left({\bf L}{\bf u}_{i}{\bf t}^{T} \right) \\ 
+ s^{2}{\bf u}_{i}^{T}{\bf u}_{i} - 2{\bf t}^{T}{\bf v}_{k} + {\bf t}^{T}{\bf t} ]
\end{multline}
and our resulting optimization problem is

\begin{equation*}
\begin{aligned}
& \underset{ {\bf L}, {\bf t}, s }{\text{minimize}}
& & e({\bf U},{\bf V};{\bf L},{\bf t}, s ) \\
& \text{subject to}
& & {\bf L}{\bf L}^{T} = {\bf I}_{N \times N}, \\
&&& \text{det} \; {\bf L} = 1.
\end{aligned}
\end{equation*}

The Lagrangian for the above (constrained) optimization problem is 

\begin{equation}
\label{eq:Lagrange_s}
\mathcal{L}({\bf L}, {\bf t}, s, \boldsymbol{\alpha}, \lambda) = e({\bf U},{\bf V};{\bf L},{\bf t}, s ) + \text{Tr} \left( \boldsymbol{\alpha}({\bf L}{\bf L}^{T} - {\bf I}_{N \times N}) \right) + \lambda \left( \text{det} \; {\bf L} - 1 \right)
\end{equation}
Optimizing Eq. (\ref{eq:Lagrange_s}) with respect to ${\bf t}$ yields 

\begin{equation}
\label{eq:trans_s}
\hat{\boldsymbol{t}} = \overline{\boldsymbol{v}} - s \boldsymbol{L}\overline{\boldsymbol{u}}
\end{equation}
Substituting Eq.(\ref{eq:trans_s}) back into our Lagrangian and then partial differentiating it with respect to ${\bf L}$ yields 

\begin{equation*}
\label{eq:pdiff_rot2}
\frac{\partial \mathcal{L}}{\partial {\bf L}} = 2\left(\boldsymbol{\alpha L} - s{\bf Z} \right) + \lambda {\bf L}= \boldsymbol{0}_{N \times N}, 
\end{equation*}
which solving for ${\bf L}$ yields

\begin{equation*}
\hat{\boldsymbol{L}} = s\left( \boldsymbol{\alpha} + \frac{\lambda}{2}{\bf I}_{N \times N} \right)^{-1}{\bf Z}
\end{equation*}
Exploiting the fact that ${\bf L}$ is an orthogonal matrix, we obtain

\begin{equation}
\label{eq:opt_rot2}
\hat{\boldsymbol{L}} = \left( \sqrt{ \boldsymbol{Z}\boldsymbol{Z}^{T} } \right)_{\textit{PD}}^{-1} \boldsymbol{Z},
\end{equation}
which is the same rotation matrix we obtained in the non-scalar case. Thus, the inclusion of a fixed scale into our affine transform does not affect the rotation matrix. 

Substituting both Eqs. (\ref{eq:trans_s}) and (\ref{eq:opt_rot2}) in Eq. (\ref{eq:Lagrange_s}) results in $\mathcal{L}({\bf L}, {\bf t}, s, \boldsymbol{\alpha}, \lambda) = \mathcal{L}(s, \boldsymbol{\alpha}, \lambda)$. Optimizing the Lagrangian with respect to $s$ yields

\begin{equation}
\label{eq:pdiff_s}
\frac{\partial \mathcal{L}}{\partial s} = 2 \left( s\left(  \sum_{i=1}^{N_{U}}\sum_{k=1}^{N_{V}}m_{ik}{\bf u}_{i}^{T} {\bf u}_{i} -  \overline{\boldsymbol{u}}^{T} \; \overline{\boldsymbol{u}} \right) - \text{Tr}\left( \left(\sqrt{ \boldsymbol{Z}\boldsymbol{Z}^{T} }\right)_{\textit{PD}}   \right) \right) = 0
\end{equation}
Note that 

\begin{equation}
\sum_{i=1}^{N_{U}}\sum_{k=1}^{N_{V}}m_{ik}{\bf u}_{i}^{T} {\bf u}_{i} -  \overline{\boldsymbol{u}}^{T} \; \overline{\boldsymbol{u}} = \sum_{j=1}^{N} \sigma_{u^{j}}^{2},
\end{equation}
where 

\begin{equation*}
\sigma_{u^{j}}^{2} = \sum_{i=1}^{N_{U}}\sum_{k=1}^{N_{V}}m_{ik}\left(u_{i}^{j} \right)^{2} - \left( \sum_{i=1}^{N_{U}}\sum_{k=1}^{N_{V}}m_{ik} u_{i}^{j} \right)^{2}
\end{equation*}
In other words, $\sigma_{u^{j}}^{2}$ is the variance of feature $u^{j}$. Solving Eq. (\ref{eq:pdiff_s}) for $s$ yields

\begin{equation}
\label{eq:scaleeee}
\hat{s} = \frac{\text{Tr}\left( \left(\sqrt{\boldsymbol{Z}\boldsymbol{Z}^{T}}\right)_{\textit{PD}} \right) }{\sum_{j=1}^{N} \sigma_{u^{j}}^{2}}
\end{equation}
Since both the numerator and denominator of Eq. (\ref{eq:scaleeee}) are always positive, it is always guaranteed that $\hat{s} > 0$; this is reasonable because a negative scale defies physical interpretation.  

So the optimal alignment parameters are

\begin{empheq}[box=\fbox]{align}
\hat{\boldsymbol{L}} = \left( \sqrt{ \boldsymbol{Z}\boldsymbol{Z}^{T} } \right)_{\textit{PD}}^{-1} \boldsymbol{Z} \nonumber \\
\hat{s} = \frac{\text{Tr}\left( \left(\sqrt{\boldsymbol{Z}\boldsymbol{Z}^{T}}\right)_{\textit{PD}} \right) }{\sum_{j=1}^{N} \sigma_{u^{j}}^{2}} \nonumber \\
\hat{\boldsymbol{t}} = \overline{\boldsymbol{v}} - \frac{\text{Tr}\left( \left(\sqrt{\boldsymbol{Z}\boldsymbol{Z}^{T}}\right)_{\textit{PD}} \right) \left( \sqrt{ \boldsymbol{Z}\boldsymbol{Z}^{T} } \right)_{\textit{PD}}^{-1} \boldsymbol{Z} \; \overline{\boldsymbol{u}} }{\sum_{j=1}^{N} \sigma_{u^{j}}^{2}}  \nonumber 
\end{empheq}
Or equivalently,

\begin{empheq}[box=\fbox]{align}
\hat{\boldsymbol{L}} = {\bf P} \left(\sqrt{{\bf D}}\right)_{\textit{PD}}^{-1} {\bf P}^{T} \boldsymbol{Z} \nonumber \\
\hat{s} = \frac{\text{Tr}\left( {\bf P} \left(\sqrt{{\bf D}}\right)_{\textit{PD}} {\bf P}^{T} \right) }{\sum_{j=1}^{N} \sigma_{u^{j}}^{2}} \nonumber \\
\hat{\boldsymbol{t}} = \overline{\boldsymbol{v}} - \frac{\text{Tr}\left( {\bf P} \left(\sqrt{{\bf D}}\right)_{\textit{PD}} {\bf P}^{T} \right) {\bf P} \left(\sqrt{{\bf D}}\right)_{\textit{PD}}^{-1} {\bf P}^{T} \boldsymbol{Z} \; \overline{\boldsymbol{u}} }{\sum_{j=1}^{N} \sigma_{u^{j}}^{2}}  \nonumber 
\end{empheq}

The minimum squared error is found by substituting the optimal alignment parameters back into Eq. (\ref{eq:dist_s}), which yields 

\begin{equation}
\begin{split}
e_{\text{min}}  &= \sum_{j=1}^{N} \sigma_{v^{j}}^{2}  - \frac{ \left[ \text{Tr} \left( \left(\sqrt{\boldsymbol{Z}\boldsymbol{Z}^{T}}\right)_{\textit{PD}} \; \right) \right]^{2} }{\sum_{j=1}^{N} \sigma_{u^{j}}^{2}}  \\
& = \sum_{j=1}^{N} \sigma_{v^{j}}^{2}  - \frac{ \left[ \text{Tr} \left( {\bf P} \left(\sqrt{{\bf D}}\right)_{\textit{PD}} {\bf P}^{T} \right) \right]^{2} }{\sum_{j=1}^{N} \sigma_{u^{j}}^{2}}
\end{split}
\end{equation}

\subsection{Uncoupled Weights: An Ill-Posed Problem }
We now examine a specific case that makes the minimization problem ill-posed, i.e. no solution exists. Assume that $m_{ik}$ is separable, i.e. $m_{ik}=\sigma_{i}\gamma_{k}$, which means that  

\begin{eqnarray*}
\overline{\boldsymbol{u}}=\sum_{i=1}^{N_{U}}\sigma_{i}{\bf u}_{i} \\
\overline{\boldsymbol{v}}=\sum_{k=1}^{N_{V}}\gamma_{k}{\bf v}_{k}
\end{eqnarray*}
Consequently, the cross-covariance matrix becomes

\begin{equation*}
{\bf Z} = \sum_{k=1}^{N_{V}}\gamma_{k}{\bf v}_{k}\sum_{i=1}^{N_{U}}\sigma_{i} {\bf u}_{i}^{T} - \overline{\boldsymbol{v}} \; \overline{\boldsymbol{u}}^{T} = \boldsymbol{0}_{N \times N}
\end{equation*}
As a result, an infinite number of rotation matrices are admissible. Thus, a unique solution for the minimization problem is guaranteed if and only if the weight term is coupled between the two point-sets. A special sub-case is when the weight term is fixed for all possible pairings, i.e. $m_{ik}=c$ for all $i$ and $k$, where $c$ is a real constant.

\subsection{Relation to Labeled Scenario}
In the case where the correspondence is known \textit{a priori}, we have $m_{ik}=w_{i}\gamma_{ik}$, where

\begin{align*}
  \gamma_{ik} = \left\{\def\arraystretch{1.2}%
  \begin{array}{@{}c@{\quad}l@{}}
   0, &  \text{points $i$ and $k$ do not correspond} \\
    1, & \text{points $i$ and $k$ do correspond} \\
  \end{array}\right.
\end{align*}
So our cost function reduces to

\begin{equation}
\label{eq:dist_labeled}
D({\bf U},{\bf V};a, b, s_{x}, s_{y}, \theta)=\frac{\sum_{n=1}^{N} w_{n}({\bf u}_{n}^{'}-{\bf v}_{n})^{T}({\bf u}_{n}^{'}-{\bf v}_{k})}{\sum_{n=1}^{N} w_{i}},
\end{equation}
where $N \leq \min \{N_{U},N_{V} \}$ is the number of matching pairs between the two sets and $w_{n}$ is the "strength" of the match between the two points forming pair $n$. The solution to Eq. (\ref{eq:dist_labeled}) is the same as that derived in unlabeled scenario, except that now the (weighted) averages, variances, and covariances are no longer coupled. Eq. (\ref{eq:dist_labeled}) is the cost function minimized in \cite{kabsch.1976,arun.1987,umeyama.1991,chang.1997}.

\section{Numerical Implementation}

We have derived the closed-form solutions for the optimal alignment parameters for two different cases: the absence of scale and the presence of a uniform scale. Without loss of generality, the present discussion will focus on the case of uniform scale. 
 
Denote ${\bf M}$ as the data structure (say an array) storing all $N_{U}N_{V}$ potential pairings. After alignment, the Euclidean distance between any two (cross) points forming a pair is

\begin{equation*}
\Delta_{ik} = \sqrt{ (\hat{s} \hat{\boldsymbol{L}}{\bf u}_{i} + \hat{\boldsymbol{t}} - {\bf v}_{k})^{T} (\hat{s} \hat{\boldsymbol{L}}{\bf u}_{i} + \hat{\boldsymbol{t}} - {\bf v}_{k}) }
\end{equation*}
If a pair constitutes a genuine match, then ideally $\Delta_{ik}$ will be small, and we will want to keep it. And if the pair is spurious, then $\Delta_{ik}$ will be large, and we will want to discard it. To do that, we need to compare each pair's $\Delta_{ik}$ to some threshold, $T$. If for a given pair,


\begin{equation*}
\Delta_{ik} > T,
\end{equation*}
then the pair is an outlier and we remove it from the array ${\bf M}$. Otherwise, we recompute its weight as 

\begin{equation}
m_{ik}=1-\frac{\Delta_{ik}}{T}
\end{equation}

Upon iterating across every pair in ${\bf M}$, we count the number of pairs left in the array. If no pairs have been removed, this is indicative of the threshold being too large, so we repeat Step 2 using the threshold $T := T - \epsilon$, where $0 < \epsilon < T$. 

If pairs have been removed, then we need to check if the convergence criterion has been met. We define convergence as when the number of pairs does not exceed the maximum number of allowable one-to-one matches, i.e. $\text{length}({\bf M}) < \min{\left( N_{U}, N_{V} \right)}$. If this happens to be the case, then we are done; the remaining pairs in ${\bf M}$ form the optimal matching pairs. Otherwise, we repeat Stage II using the updated ${\bf M}$ and new weights. 

Note that the algorithm consists of two hyper-parameters: the max threshold, $T$, and the threshold spacing, $\epsilon$. They have to be tuned appropriately with respect to the data. 

We now summarize the proposed algorithm as a pseudocode:

\begin{algorithmic}[1]
\WHILE {number of pairs $> \min(N_{U},N_{V})$ and $T > 0$} \STATE{Align query object to template
\FOR {each pair $M_{j}$ in array {\bf \textit{M}}} \STATE{compute weighted sum, $\Delta_{j}$, of radial displacements
\IF {$\Delta_{j} > T$} \STATE{ remove pair $M_{j}$ from array {\bf \textit{M}}} 
\ELSE \STATE{compute new weight of $M_{j}$} 
\ENDIF 
}
\ENDFOR
\IF {no pairs are removed} \STATE{$T := T-\epsilon$} 
\ENDIF
}
\ENDWHILE
\end{algorithmic}

\subsection{Computational Complexity}
Denote $I(T, \epsilon)$ as the number of iterations until convergence is reached; it is a function of the two hyper-paramters. The computational complexity of the proposed algorithm is then $\mathcal{O}(I N_{U} N_{V})$. In the best case scenario, the algorithm will achieve convergence after a single iteration, i.e. $I = 1$. 

Let us look at the scenario where no pairs will be removed during each iteration of alignment. Such a scenario is the worst-case from a practical standpoint, but not necessarily from an algorithmic (i.e. computational complexity) perspective. In such a scenario, the total number of attempted alignments is $I(T, \epsilon) = \left \lceil{\frac{T}{\epsilon}}\right \rceil$, so the complexity is $\mathcal{O}(\left \lceil{\frac{T}{\epsilon}}\right \rceil N_{U}N_{V})$. Note, a worst computational complexity than this could be achieved; for example, having more than $\left \lceil{\frac{T}{\epsilon}}\right \rceil$ iterations until convergence is reached is completely plausible because any given threshold may be used more than once. 

\subsection{Similarity Score}
Based on the set of (optimal) matching point pairs outputted by the proposed algorithm, a similarity score measuring the strength of the match between the two point-sets can be computed from the minimum squared error generated by the optimal alignment parameters, which recall for the case of uniform scale is 

\begin{equation*}
e_{\text{min}}  =  \sum_{j=1}^{N} \sigma_{v^{j}}^{2}  - \frac{ \left[ \text{Tr} \left( {\bf P} \left(\sqrt{{\bf D}}\right)_{\textit{PD}} {\bf P}^{T} \right) \right]^{2} }{\sum_{j=1}^{N} \sigma_{u^{j}}^{2}}
\end{equation*}
It would ideally be small for genuine matches and large for dissimilar point-sets. 

\section{Discussion}
An unsupervised, iterative $N$-dimensional point-set registration algorithm for unlabeled data (i.e. correspondence between points is unknown) and based on linear least squares has been proposed. The algorithm considers all possible point pairings and iteratively aligns the two sets until the number of point pairs does not exceed the maximum number of allowable one-to-one pairings. 

Note that the output the of algorithm, i.e. the optimal matching point-pairs, may not necessarily be injective. In fact, a point in ${\bf U}$ may match to more than one point in ${\bf V}$, or more than one point in ${\bf U}$ may match to the same point in ${\bf V}$. Such a situation frequently arises in fingerprint matching where due to image noise and/or faulty image processing, a minutia in the query image may genuinely correspond to more than one minutia in the reference image. 

The proposed algorithm may be utilized in a wide variety of matching problems, beyond just fingerprint matching. For example, assuming ${\bf U}$ and ${\bf V}$ describe two different sets of individuals, we may be interested in identifying those pairs of individuals that are most likely to become friends. Another example is where ${\bf U}$ describing a set of individuals and ${\bf V}$ a set of companies, and we seek to find which company a given individual best matches to. In these two examples, injectivity may not be a realistic assumption; there may be more than one company that a given individual would fit well in, or there may be one company where more than one individual would be a great fit.

%

\bibliographystyle{IEEEtran}
\bibliography{refs}

\end{document}